\documentclass{article}

\usepackage{PRIMEarxiv}

\usepackage[utf8]{inputenc} 
\usepackage[T1]{fontenc}    
\usepackage{hyperref}       
\usepackage{url}            
\usepackage{booktabs}       
\usepackage{amsfonts}       
\usepackage{nicefrac}       
\usepackage{microtype}      
\usepackage{lipsum}
\usepackage{fancyhdr}       
\usepackage{graphicx}       
\usepackage{natbib}
\usepackage{subcaption}

\graphicspath{{media/}}     

\title{Vision Transformers Need Better Token Interaction
}

\author{
  Su Linxiang \\
  University of Szeged \\
  \texttt{su\_linxiang@126.com} \\
}

\begin{document}
\maketitle

\begin{abstract}
Vision Transformers (ViTs) can learn strong image-level representations while their patch representations become less effective for dense prediction during prolonged training. 
We revisit this dense degradation phenomenon and argue that it is not fully explained by high-norm artifacts alone. 
Instead, we characterize \emph{semantic diffusion}: an optimization shortcut in which global semantic information spreads through patch tokens beyond what is locally justified. 
Our analysis shows that dense representation quality is not captured by locality alone: shallow features can remain better aligned with foreground regions yet underperform deeper features, and \texttt{[CLS]} features remain complementary for dense prediction. 
These observations suggest that the goal should not be to remove global context, but to make token interactions more selective. 
We therefore study sparse attention as a minimal intervention, replacing softmax attention with entmax-1.5 while preserving global token connectivity. 
On DINOv1 ViT-S/16 trained for 200 epochs on ImageNet-1K, this change preserves ImageNet linear probing accuracy and substantially improves semantic segmentation performance: VOC mIoU increases from 42.80 to 48.78, ADE20K from 19.85 to 21.97, and Cityscapes from 36.79 to 37.87. 
These results suggest that selective token mixing is a simple and effective bias for improving dense ViT representations.
\end{abstract}


\begin{figure}[htbp]
    \centering
    \includegraphics[width=1\linewidth]{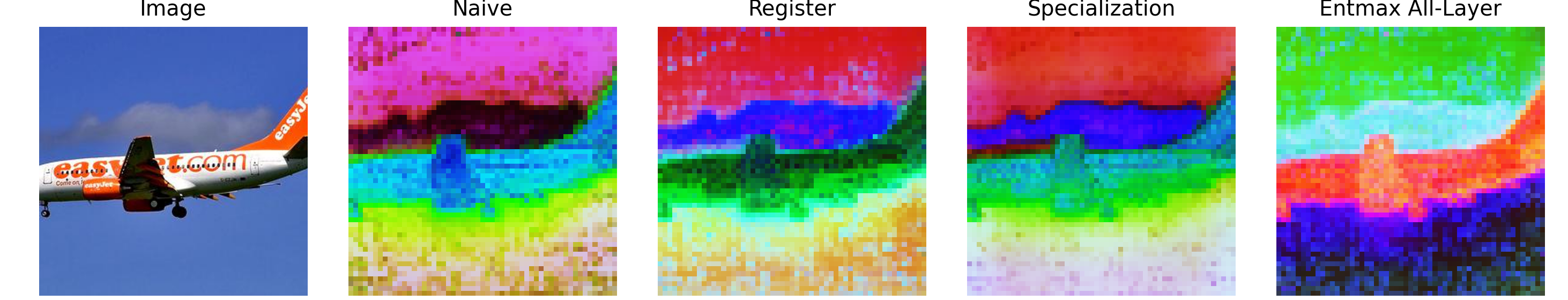}
    \caption{Visualization of the impact of our proposed sparse attention, compared with other baselines. We display the first PCA components of model outputs in RGB.}
    \label{fig:pca}
\end{figure}

\section{Introduction}

Vision Transformers (ViTs) have become a standard architecture for learning transferable visual representations \citep{dosovitskiy2021image,caron2021emerging,oquab2023dinov2}. Beyond strong image-level recognition performance, a desirable ViT representation should also preserve spatially meaningful patch features that transfer effectively to dense prediction tasks such as semantic segmentation and monocular depth estimation.

However, recent studies suggest that these objectives can diverge during training. As training progresses, image-level performance may continue improving while dense representations gradually become noisier and less effective for downstream dense tasks. This phenomenon has been associated with artifacts or outlier tokens in ViTs \citep{darcet2024vision,shi2026more}. In this work, we argue that such artifacts are likely a symptom rather than the root cause. Instead, we hypothesize that the main issue is an optimization shortcut in which global semantic information diffuses into patch tokens beyond what is locally justified, a phenomenon we term \emph{semantic diffusion}.

LaSt-ViT attributes dense degradation to the combination of coarse image-level supervision and unrestricted global token interactions, which encourages background patches to participate in representing global semantics \citep{shi2026more}. Other approaches attempt to mitigate this issue through register tokens, token specialization, attention modifications, or dense relational supervision \citep{darcet2024vision,marouani2026revisiting,oquab2023dinov2, simeoni2025dinov3}. While effective, these methods do not directly address the shortcut itself. Register tokens mainly provide dedicated storage for leaked information, heuristic token filtering relies on manually designed criteria, and teacher-based dense constraints require an additional dense-good teacher model.

Our starting point is that the objective should not be to eliminate global context or suppress background tokens altogether. Background regions often contain useful contextual cues, and globally shared semantics can still benefit downstream tasks. Instead, the goal should be to encourage \emph{useful token interactions}: interactions that preserve local structure while selectively incorporating global contextual information.

Motivated by this perspective, we study sparse attention as a simple and direct intervention on attention geometry. Rather than restricting receptive fields through local windows \citep{liu2021swintransformerhierarchicalvision} or introducing additional register tokens \citep{darcet2024vision}, we retain global connectivity but replace dense softmax attention with entmax-1.5 \citep{peters2019sparse,correia2019adaptively}. This produces sparse and selective attention distributions through only a minimal architectural modification.

\section{Related Work}

\paragraph{Artifacts in ViTs.}
ViT patch features can exhibit high-norm outliers, noisy PCA visualizations, or degraded dense-task performance despite strong image-level recognition ability. DINOv2 popularized register tokens as a practical mechanism for improving dense representations by introducing additional storage slots \citep{darcet2024vision,oquab2023dinov2}. 

Recent work further suggests that such artifacts may emerge systematically from attention dynamics rather than random optimization failures. In large language models, systematic outliers have been linked to the softmax attention mechanism itself, where they act as implicit context-aware scaling factors \citep{an2025systematicoutlierslargelanguage}. In vision transformers, LAST-ViT attributes artifacts to a lazy aggregation behavior, where semantically irrelevant background patches are exploited as shortcuts for representing global semantics \citep{shi2026more}. 

Our work is aligned with these perspectives. However, instead of managing artifacts after they emerge, we focus on modifying the token interaction mechanism itself to make semantic diffusion less favorable during optimization.


\paragraph{Dense Performance Degradation.}
Recent studies observe that prolonged self-supervised training can continuously improve global recognition performance while simultaneously degrading dense representations. This phenomenon was first implicitly observed in DINOv2 \citep{oquab2023dinov2} and later studied more explicitly in DINOv3 \citep{simeoni2025dinov3}. DINOv3 proposes Gram anchoring, where an earlier checkpoint with stronger dense performance is used as a teacher to regularize later training stages, achieving strong empirical improvements. In contrast, we focus on understanding and mitigating the underlying optimization behavior itself, which we term \emph{semantic diffusion}. Importantly, our approach is orthogonal to Gram anchoring and can potentially improve the quality of the teacher representations as well.

\paragraph{CLS--Patch Interaction.}
\label{sec:specialization}
The CLS token and patch tokens play inherently different roles but are typically processed through the same transformer blocks. Recent work suggests that ViTs implicitly attempt to separate these token types and that explicit CLS/patch specialization can improve dense representations \citep{marouani2026revisiting}. Our findings are consistent with this perspective: modifying CLS--patch interaction appears more effective than directly constraining final patch representations. Unlike prior specialization approaches, we intervene at the attention normalization level rather than introducing separate processing branches or additional parameters.


\paragraph{Sparse Attention.}
Sparse attention has primarily been explored for efficient sequence modeling and scalable transformers. Entmax and adaptive sparse transformers demonstrate that replacing softmax with sparse probability mappings can produce selective attention distributions while remaining fully differentiable \citep{peters2019sparse,correia2019adaptively}. In ViTs, sparsity has often been introduced for computational or architectural efficiency \citep{ibtehaz2024accvitatrousconvolutions,chen2023sparsevitrevisitingactivationsparsity}. In contrast, we use sparsity as a representational bias rather than a compute optimization. Our goal is to make semantic diffusion less favorable during optimization by encouraging more selective token interactions.


    \label{fig:voc_layercls}

\begin{figure*}[htbp]
    \centering
    
    \begin{subfigure}[c]{0.38\linewidth}
        \centering
        \includegraphics[width=\linewidth]{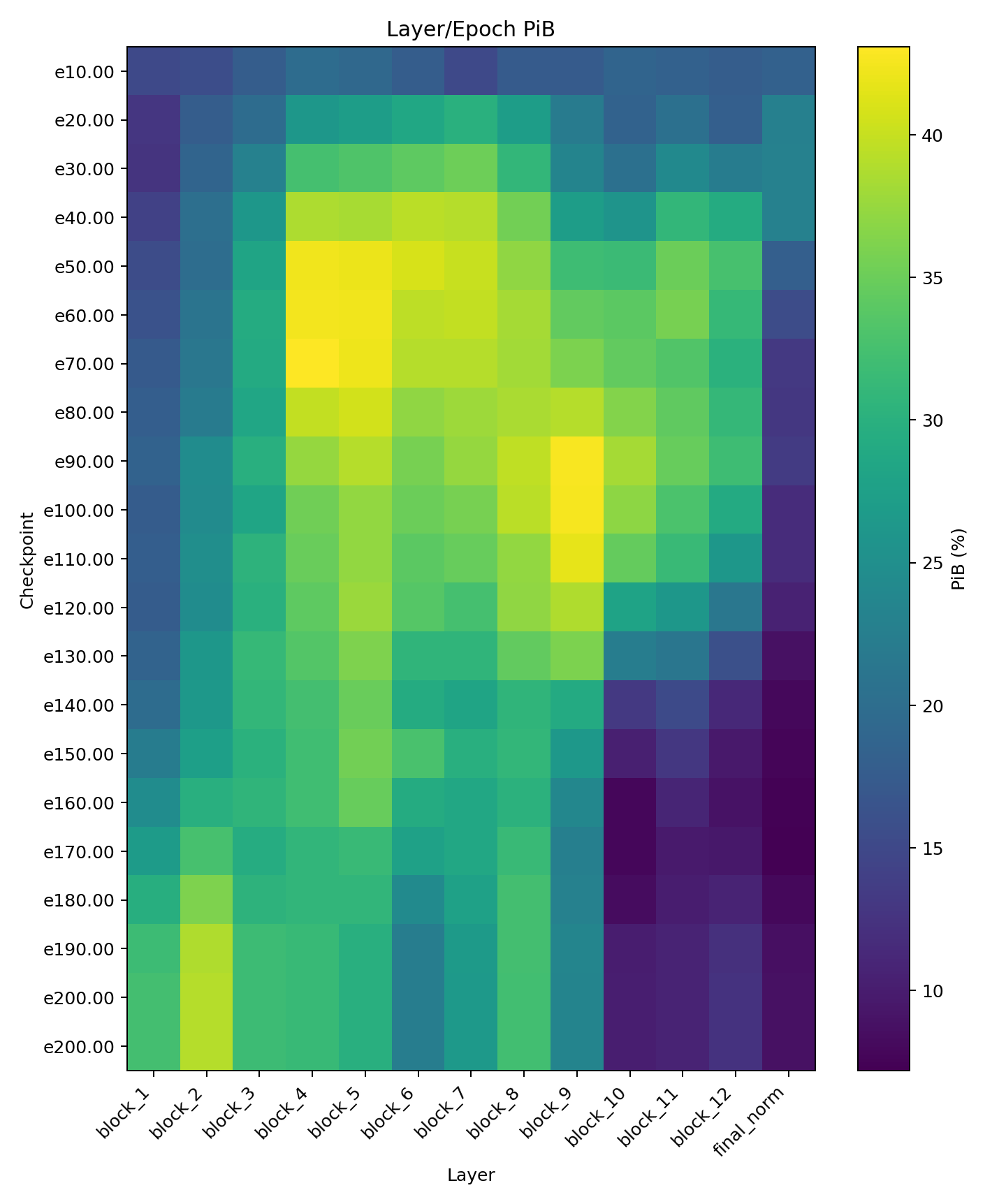}
        \caption{Layer/epoch heatmap of point-in-box (PiB). }
        \label{fig:pib}
    \end{subfigure}
    \hfill
    \begin{subfigure}[c]{0.58\linewidth}
        \centering
        \includegraphics[width=\linewidth]{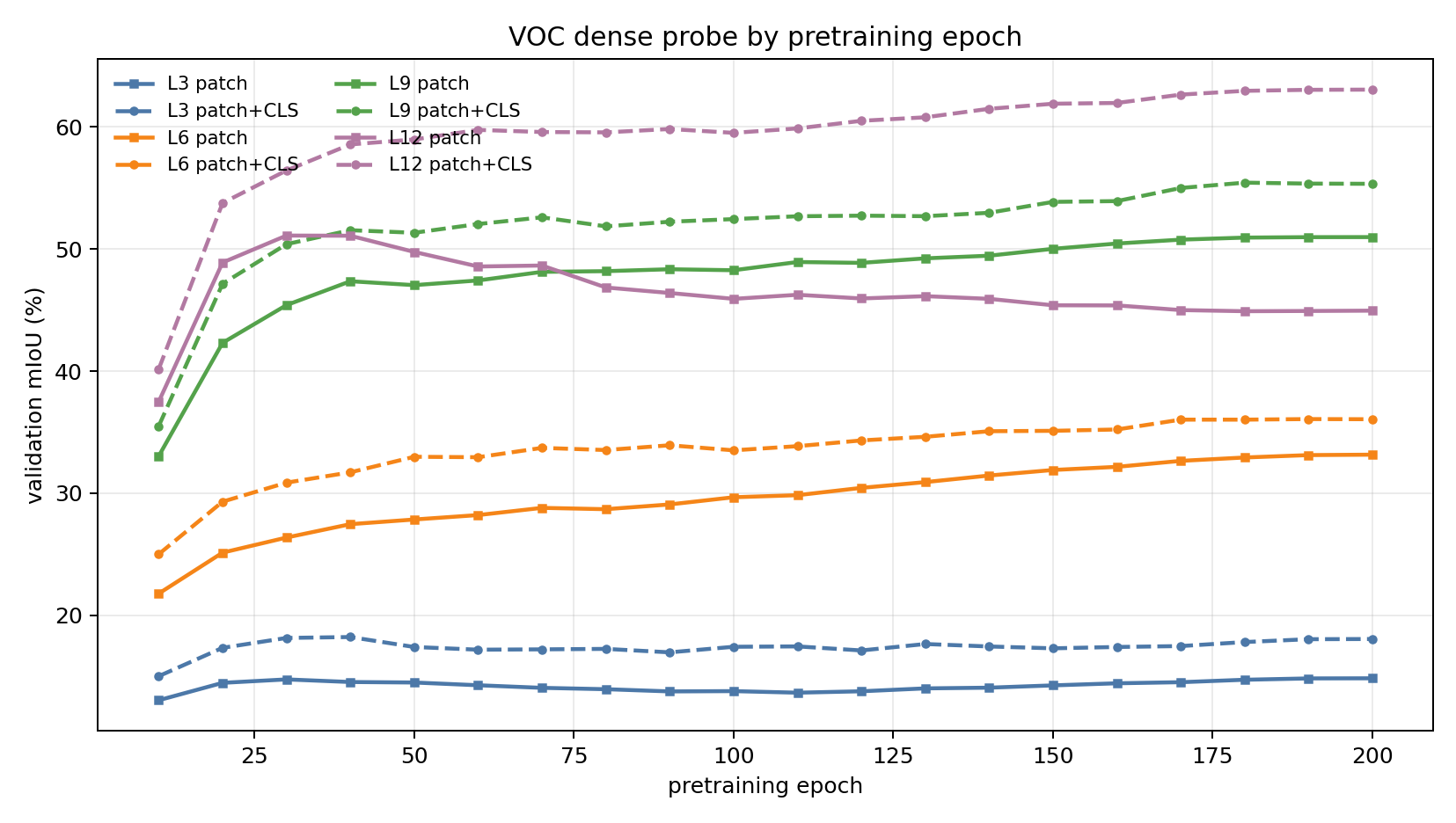}
        \caption{VOC dense-probe mIoU across pretraining epochs and feature layers.}
        \label{fig:voc_layercls}
    \end{subfigure}

    \caption{
    Relationship between CLS spatial grounding and dense prediction quality across training. 
    Left: PiB heatmap showing how well CLS-aligned patches remain localized within foreground regions. 
    Right: VOC dense probing performance across layers and epochs.}
    
    \label{fig:pib_voc_combined}
\end{figure*}

\section{Method}

\subsection{Semantic Diffusion as an Optimization Shortcut}

We hypothesize that under global training objectives, ViTs exploit unrestricted token interactions to distribute class-relevant semantics across patch tokens in a locally inconsistent manner. This optimization shortcut can improve global image-level recognition while gradually degrading patch-level semantic structure, especially for dense prediction tasks.

To study this behavior, we analyze point-in-box (PiB), defined as the proportion of images where the patch with the highest cosine similarity to the \texttt{[CLS]} token lies inside the foreground bounding box \citep{shi2026more}. Intuitively, higher PiB indicates stronger spatial alignment between \texttt{[CLS]} semantics and foreground regions.

As shown in Figure~\ref{fig:pib}, shallow layers consistently exhibit higher PiB scores during training, whereas deeper layers show progressively weaker localization behavior. However, this trend does not imply that shallow representations are more useful for dense prediction. To further investigate this, we evaluate dense representations from different transformer layers on VOC semantic segmentation, optionally concatenating patch features with the corresponding \texttt{[CLS]} feature.

Figure~\ref{fig:voc_layercls} reveals two important observations. First, despite stronger PiB scores, shallow-layer features perform substantially worse on dense tasks. This suggests that locality alone is insufficient for high-quality dense representations; semantic richness is also necessary. Second, concatenating the \texttt{[CLS]} feature consistently improves dense performance and largely mitigates the degradation observed in the final transformer layers. This indicates that although semantic information diffuses through patch tokens during training \citep{simeoni2025dinov3}, the \texttt{[CLS]} representation itself still contains complementary global information useful for downstream dense prediction.

These observations suggest that an effective intervention should not simply remove global context, for example through local window attention \citep{liu2021swintransformerhierarchicalvision}. Likewise, enforcing hand-designed interaction rules, such as manually filtering stable tokens for CLS aggregation \citep{shi2026more}, may overly constrain optimization behavior. Instead, the goal should be to preserve global receptive fields while encouraging more selective and meaningful token interactions.

Motivated by this perspective, we explore sparse attention as a simple and direct intervention on token interaction geometry.

\subsection{Sparse Attention}

Given queries \(Q\), keys \(K\), and values \(V\), standard attention computes
\[
A = \mathrm{softmax}(QK^\top / \sqrt{d}), \quad Y = AV.
\]

We replace the softmax normalization with entmax-1.5:
\[
A = \mathrm{entmax}_{1.5}(QK^\top / \sqrt{d}), \quad Y = AV.
\]

For an input score vector \(z \in \mathbb{R}^n\), entmax-1.5 is defined as
\[
\mathrm{entmax}_{1.5}(z)_i
=
\left[
\frac{z_i - \tau(z)}{2}
\right]_+^2,
\]
where \([x]_+ = \max(x,0)\), and the threshold \(\tau(z)\) is chosen such that
\[
\sum_{i=1}^n \mathrm{entmax}_{1.5}(z)_i = 1.
\]

Unlike windowed attention, sparse attention preserves the full global candidate set for every token. Unlike register tokens, it introduces no additional storage tokens. Unlike heuristic filtering approaches such as LaSt-ViT, it does not rely on manually designed token selection rules based on low-pass stability estimates. Instead, the intervention only changes how attention mass is distributed: low-utility interactions can receive exact zero weight, encouraging more selective token communication and making semantic diffusion less favorable during optimization.

In our experiments, we adopt the simplest possible setting and replace softmax with entmax-1.5 in all transformer layers without introducing any additional architectural modification or artifact-handling mechanism. 

\section{Experiments}

\begin{table*}[t]
\centering
\small
\begin{tabular}{lrrrrrrr}
\hline
Method & ImageNet LP & VOC & ADE20K & Cityscapes & NYUv2 & SUN RGB-D & KITTI \\
 & Top-1 $\uparrow$ & mIoU $\uparrow$ & mIoU $\uparrow$ & mIoU $\uparrow$ & RMSE $\downarrow$ & RMSE $\downarrow$ & RMSE $\downarrow$ \\
\hline
CLS baseline & 69.50 & 42.80 & 19.85 & 36.79 & 0.834 & 0.674 & 6.743 \\
Register-4 & 70.03 & 44.34 & 19.52 & 36.00 & \textbf{0.814} & 0.667 & 6.749 \\
Specialization & 70.04 & 46.22 & 20.51 & 36.45 & 0.829 & \textbf{0.660} & \textbf{6.723} \\
Entmax-1.5 & \textbf{70.06} & \textbf{48.78} & \textbf{21.97} & \textbf{37.87} & 0.829 & 0.669 & 6.739 \\
\hline
\end{tabular}
\caption{Evaluation of DINOv1 ViT-S/16 models trained for 200 epochs on ImageNet-1K. Dense metrics report the best validation performance during probe training.}
\label{tab:nococo}
\end{table*}

\subsection{Experimental Settings}

We evaluate DINOv1 \citep{caron2021emerging} ViT-S/16 models trained for 200 epochs on ImageNet-1K \citep{imagenet15russakovsky}. 
All methods are evaluated using the same training and probing protocols. 
Following the evaluation protocol of DINOv2 \citep{oquab2023dinov2}, we consider both global image-level recognition and dense prediction benchmarks.

\paragraph{Compared methods.}
We compare four variants. 
The \emph{CLS baseline} is the standard DINOv1 ViT-S/16 model with softmax attention. 
\emph{Register-4} adds four register tokens following prior work on improving ViT dense features \citep{darcet2024vision,oquab2023dinov2}. 
\emph{Specialization} introduces explicit CLS/patch specialization to encourage different processing for the \texttt{[CLS]} token and patch tokens \citep{marouani2026revisiting}. 
\emph{Entmax-1.5} replaces softmax attention with entmax-1.5 in all transformer layers, without adding extra tokens, separate processing branches, or additional dense supervision.

\paragraph{Image-level recognition.}
For the global task, we perform linear probing on ImageNet classification. 
A linear classifier is trained on top of the frozen \texttt{[CLS]} feature using SGD for 12,500 iterations with random-resized-crop augmentation. 
We perform a learning-rate grid search over 
\(\{10^{-5}, 2{\times}10^{-5}, 5{\times}10^{-5}, 10^{-4}, 2{\times}10^{-4}, 5{\times}10^{-4}, 10^{-3}, 2{\times}10^{-3}, 5{\times}10^{-3}, 10^{-2}, 2{\times}10^{-2}, 5{\times}10^{-2}, 10^{-1}\}\), 
and report the best validation top-1 accuracy.

\paragraph{Semantic segmentation.}
For dense semantic prediction, we evaluate on VOC \citep{pascal-voc-2012}, ADE20K \citep{8100027,zhou2018semantic}, and Cityscapes \citep{Cordts2016Cityscapes}. 
We train a linear segmentation head for 40,000 iterations with learning rate \(10^{-3}\). 
The head is applied to frozen patch features after the final layer normalization, followed by a trained batch normalization layer. 
We report mean Intersection-over-Union (mIoU), where higher is better.

\paragraph{Depth estimation.}
For monocular depth estimation, we evaluate on NYUv2 \citep{Silberman:ECCV12}, SUN RGB-D \citep{Song_2015_CVPR}, and KITTI \citep{Geiger2013IJRR}. 
We train a linear depth probe for 38,400 iterations with learning rate \(10^{-3}\). 
Following the DINOv2 protocol, the probe input consists of patch and \texttt{[CLS]} features from four evenly spaced transformer layers of the frozen backbone, without applying the final layer normalization. 
We report root mean squared error (RMSE), where lower is better.

\subsection{Quantitative Results}

Table~\ref{tab:nococo} shows that replacing softmax attention with entmax-1.5 improves dense representations without sacrificing global recognition performance. 
Entmax-1.5 achieves the best ImageNet linear probing accuracy among the evaluated methods, improving over the CLS baseline from 69.50\% to 70.06\%.

The gains are most pronounced for semantic segmentation. 
Compared with the CLS baseline, Entmax-1.5 improves VOC by 5.98 mIoU, ADE20K by 2.12 mIoU, and Cityscapes by 1.08 mIoU. 
It also outperforms both Register-4 and CLS/patch specialization on all three segmentation benchmarks. 
This suggests that sparse attention improves the spatial quality of patch representations while preserving the semantic information required for image-level recognition.

Depth estimation results are more moderate. 
Entmax-1.5 still improves over the CLS baseline on all three depth benchmarks, reducing RMSE by 0.005 on NYUv2, 0.005 on SUN RGB-D, and 0.004 on KITTI. 
However, it does not consistently outperform specialized baselines such as Register-4 or CLS/patch specialization. 
One possible reason is that our depth probing protocol already uses patch features from four transformer layers concatenated with the \texttt{[CLS]} feature. 
As discussed in Section~\ref{sec:specialization}, \texttt{[CLS]} and patch tokens tend to contain complementary information; concatenating them may therefore partially compensate for semantic diffusion even in the baseline model. 
Under this relatively strong probing setup, sparse attention still provides consistent improvements over the naive CLS baseline and remains competitive with more specialized methods.

Overall, the quantitative results support our main hypothesis: encouraging selective token interactions can improve dense representations, especially for semantic segmentation, while maintaining or slightly improving global image recognition.

\subsection{Qualitative Results}

For qualitative evaluation, we visualize patch features using PCA in Figure~\ref{fig:pca}. 
Compared with the CLS baseline, Register-4, and CLS/patch specialization, Entmax-1.5 produces cleaner and more spatially coherent feature maps. 
Foreground regions are more clearly separated from background regions, and object boundaries are better preserved. 
These visualizations are consistent with the quantitative segmentation results and suggest that sparse attention reduces noisy semantic diffusion across unrelated patches.

\section{Discussion}

Entmax-1.5 is not intended as a definitive solution for token interaction. 
Rather, it serves as a compact instantiation of a broader design principle: harmful semantic diffusion can be reduced by making attention mixing more selective while preserving global connectivity. 
Future work could explore more flexible sparsification mechanisms, such as layer-adaptive or input-dependent attention sparsity \citep{qiu2025gatedattentionlargelanguage}, or combine sparse attention with potentially complementary interaction mechanisms such as explicit CLS/patch specialization \citep{marouani2026revisiting}. 
Nevertheless, our results show that even a lightweight change to attention normalization can improve dense representations while maintaining global recognition performance.

\textbf{Limitations.} This study is currently limited to DINOv1 ViT-S/16 trained for 200 epochs on ImageNet-1K. We have not yet evaluated larger backbones, longer pretraining schedules, or multi-seed stability. The gains on depth estimation are modest compared with semantic segmentation, and entmax attention may introduce implementation-dependent runtime overhead (although there are highly efficient implementations \citep{goncalves2025adasplash,goncalves2026adasplash}). We therefore view sparse attention as evidence for the broader principle of selective token interaction, rather than as a final architecture design.

\section{Conclusion}

We revisit the dense degradation phenomenon in Vision Transformers and provide evidence that semantic diffusion is an important optimization shortcut behind this behavior. 
Motivated by this perspective, we propose sparse attention as a simple intervention that makes token interactions more selective without removing the global receptive field. 
By replacing softmax with entmax-1.5, our method introduces no additional labels, storage tokens, or teacher models. 
Experiments on DINOv1 ViT-S/16 show that this minimal modification improves dense prediction performance, especially for semantic segmentation, while preserving image-level recognition accuracy.


\bibliographystyle{unsrt}  
\bibliography{references}

@inproceedings{dosovitskiy2021image,
  title = {An Image Is Worth 16x16 Words: Transformers for Image Recognition at Scale},
  author = {Dosovitskiy, Alexey and Beyer, Lucas and Kolesnikov, Alexander and Weissenborn, Dirk and Zhai, Xiaohua and Unterthiner, Thomas and Dehghani, Mostafa and Minderer, Matthias and Heigold, Georg and Gelly, Sylvain and Uszkoreit, Jakob and Houlsby, Neil},
  booktitle = {International Conference on Learning Representations},
  year = {2021}
}

@inproceedings{caron2021emerging,
  title = {Emerging Properties in Self-Supervised Vision Transformers},
  author = {Caron, Mathilde and Touvron, Hugo and Misra, Ishan and J{\'e}gou, Herv{\'e} and Mairal, Julien and Bojanowski, Piotr and Joulin, Armand},
  booktitle = {Proceedings of the IEEE/CVF International Conference on Computer Vision},
  year = {2021}
}

@article{oquab2023dinov2,
  title = {{DINOv2}: Learning Robust Visual Features without Supervision},
  author = {Oquab, Maxime and Darcet, Timoth{\'e}e and Moutakanni, Th{\'e}o and Vo, Huy V. and Szafraniec, Marc and Khalidov, Vasil and Fernandez, Pierre and Haziza, Daniel and Massa, Francisco and El-Nouby, Alaaeldin and Assran, Mahmoud and Ballas, Nicolas and Galuba, Wojciech and Howes, Russell and Huang, Po-Yao and Li, Shang-Wen and Misra, Ishan and Rabbat, Michael and Sharma, Vasu and Synnaeve, Gabriel and Xu, Hu and Jegou, Herve and Mairal, Julien and Labatut, Patrick and Joulin, Armand and Bojanowski, Piotr},
  journal = {Transactions on Machine Learning Research},
  year = {2024}
}

@misc{simeoni2025dinov3,
  title={{DINOv3}},
  author={Sim{\'e}oni, Oriane and Vo, Huy V. and Seitzer, Maximilian and Baldassarre, Federico and Oquab, Maxime and Jose, Cijo and Khalidov, Vasil and Szafraniec, Marc and Yi, Seungeun and Ramamonjisoa, Micha{\"e}l and Massa, Francisco and Haziza, Daniel and Wehrstedt, Luca and Wang, Jianyuan and Darcet, Timoth{\'e}e and Moutakanni, Th{\'e}o and Sentana, Leonel and Roberts, Claire and Vedaldi, Andrea and Tolan, Jamie and Brandt, John and Couprie, Camille and Mairal, Julien and J{\'e}gou, Herv{\'e} and Labatut, Patrick and Bojanowski, Piotr},
  year={2025},
  eprint={2508.10104},
  archivePrefix={arXiv},
  primaryClass={cs.CV},
  url={https://arxiv.org/abs/2508.10104},
}

@inproceedings{darcet2024vision,
  title = {Vision Transformers Need Registers},
  author = {Darcet, Timoth{\'e}e and Oquab, Maxime and Mairal, Julien and Bojanowski, Piotr},
  booktitle = {International Conference on Learning Representations},
  year = {2024}
}

@misc{shi2026more,
  title = {Vision Transformers Need More Than Registers},
  author = {Shi, Cheng and Yu, Yizhou and Yang, Sibei},
  year = {2026},
  eprint = {2602.22394},
  archivePrefix = {arXiv},
  primaryClass = {cs.CV}
}

@inproceedings{marouani2026revisiting,
  title = {Revisiting {[CLS]} and Patch Token Interaction in Vision Transformers},
  author = {Marouani, Alexis and Sim{\'e}oni, Oriane and J{\'e}gou, Herv{\'e} and Bojanowski, Piotr and Vo, Huy V.},
  booktitle = {International Conference on Learning Representations},
  year = {2026}
}

@inproceedings{peters2019sparse,
  title = {Sparse Sequence-to-Sequence Models},
  author = {Peters, Ben and Niculae, Vlad and Martins, Andr{\'e} F. T.},
  booktitle = {Proceedings of the 57th Annual Meeting of the Association for Computational Linguistics},
  year = {2019}
}

@inproceedings{correia2019adaptively,
  title = {Adaptively Sparse Transformers},
  author = {Correia, Gon{\c{c}}alo M. and Niculae, Vlad and Martins, Andr{\'e} F. T.},
  booktitle = {Proceedings of the 2019 Conference on Empirical Methods in Natural Language Processing and the 9th International Joint Conference on Natural Language Processing},
  year = {2019}
}

@misc{ibtehaz2024accvitatrousconvolutions,
      title={ACC-ViT : Atrous Convolution's Comeback in Vision Transformers}, 
      author={Nabil Ibtehaz and Ning Yan and Masood Mortazavi and Daisuke Kihara},
      year={2024},
      eprint={2403.04200},
      archivePrefix={arXiv},
      primaryClass={cs.CV},
      url={https://arxiv.org/abs/2403.04200}, 
}

@misc{chen2023sparsevitrevisitingactivationsparsity,
      title={SparseViT: Revisiting Activation Sparsity for Efficient High-Resolution Vision Transformer}, 
      author={Xuanyao Chen and Zhijian Liu and Haotian Tang and Li Yi and Hang Zhao and Song Han},
      year={2023},
      eprint={2303.17605},
      archivePrefix={arXiv},
      primaryClass={cs.CV},
      url={https://arxiv.org/abs/2303.17605}, 
}

@misc{liu2021swintransformerhierarchicalvision,
      title={Swin Transformer: Hierarchical Vision Transformer using Shifted Windows}, 
      author={Ze Liu and Yutong Lin and Yue Cao and Han Hu and Yixuan Wei and Zheng Zhang and Stephen Lin and Baining Guo},
      year={2021},
      eprint={2103.14030},
      archivePrefix={arXiv},
      primaryClass={cs.CV},
      url={https://arxiv.org/abs/2103.14030}, 
}

@misc{qiu2025gatedattentionlargelanguage,
      title={Gated Attention for Large Language Models: Non-linearity, Sparsity, and Attention-Sink-Free}, 
      author={Zihan Qiu and Zekun Wang and Bo Zheng and Zeyu Huang and Kaiyue Wen and Songlin Yang and Rui Men and Le Yu and Fei Huang and Suozhi Huang and Dayiheng Liu and Jingren Zhou and Junyang Lin},
      year={2025},
      eprint={2505.06708},
      archivePrefix={arXiv},
      primaryClass={cs.CL},
      url={https://arxiv.org/abs/2505.06708}, 
}

@misc{an2025systematicoutlierslargelanguage,
      title={Systematic Outliers in Large Language Models}, 
      author={Yongqi An and Xu Zhao and Tao Yu and Ming Tang and Jinqiao Wang},
      year={2025},
      eprint={2502.06415},
      archivePrefix={arXiv},
      primaryClass={cs.CL},
      url={https://arxiv.org/abs/2502.06415}, 
}

@inproceedings{
goncalves2025adasplash,
title={AdaSplash: Adaptive Sparse Flash Attention},
author={Nuno Gon{\c{c}}alves and Marcos V Treviso and Andre Martins},
booktitle={Forty-second International Conference on Machine Learning},
year={2025},
url={https://openreview.net/forum?id=OWIPDWhUcO}
}

@inproceedings{
goncalves2026adasplash,
title={AdaSplash-2: Faster Differentiable Sparse Attention},
author={Nuno Gon{\c{c}}alves and Hugo Pitorro and Vlad Niculae and Edoardo Ponti and Lei Li and Andre Martins and Marcos V Treviso},
booktitle={Forty-third International Conference on Machine Learning},
year={2026},
url={https://openreview.net/forum?id=7qpvff2gWI}
}

@article{imagenet15russakovsky,
    Author = {Olga Russakovsky and Jia Deng and Hao Su and Jonathan Krause and Sanjeev Satheesh and Sean Ma and Zhiheng Huang and Andrej Karpathy and Aditya Khosla and Michael Bernstein and Alexander C. Berg and Li Fei-Fei},
    Title = { {ImageNet Large Scale Visual Recognition Challenge} },
    Year = {2015},
    journal   = {International Journal of Computer Vision (IJCV)},
    doi = {10.1007/s11263-015-0816-y},
    volume={115},
    number={3},
    pages={211-252}
}

@misc{pascal-voc-2012,
	author = "Everingham, M. and Van~Gool, L. and Williams, C. K. I. and Winn, J. and Zisserman, A.",
	title = "The {PASCAL} {V}isual {O}bject {C}lasses {C}hallenge 2012 {(VOC2012)} {R}esults",
	howpublished = "http://www.pascal-network.org/challenges/VOC/voc2012/workshop/index.html"}

@inproceedings{8100027,
  title      = {Scene Parsing through ADE20K Dataset},
  author     = {Zhou, Bolei and Zhao, Hang and Puig, Xavier and Fidler, Sanja and Barriuso, Adela and Torralba, Antonio},
  year       = 2017,
  booktitle  = {2017 IEEE Conference on Computer Vision and Pattern Recognition (CVPR)},
  volume     = {},
  number     = {},
  pages      = {5122--5130},
  doi        = {10.1109/CVPR.2017.544}
}

@misc{zhou2018semantic,
  title         = {Semantic Understanding of Scenes through the ADE20K Dataset},
  author        = {Bolei Zhou and Hang Zhao and Xavier Puig and Tete Xiao and Sanja Fidler and Adela Barriuso and Antonio Torralba},
  year          = 2018,
  eprint        = {1608.05442},
  archiveprefix = {arXiv},
  primaryclass  = {cs.CV}
}

@inproceedings{Cordts2016Cityscapes,
  title={The Cityscapes Dataset for Semantic Urban Scene Understanding},
  author={Cordts, Marius and Omran, Mohamed and Ramos, Sebastian and Rehfeld, Timo and Enzweiler, Markus and Benenson, Rodrigo and Franke, Uwe and Roth, Stefan and Schiele, Bernt},
  booktitle={Proc. of the IEEE Conference on Computer Vision and Pattern Recognition (CVPR)},
  year={2016}
}

@inproceedings{Silberman:ECCV12,
  author    = {Nathan Silberman, Derek Hoiem, Pushmeet Kohli and Rob Fergus},
  title     = {Indoor Segmentation and Support Inference from RGBD Images},
  booktitle = {ECCV},
  year      = {2012}
}

@InProceedings{Song_2015_CVPR,
author = {Song, Shuran and Lichtenberg, Samuel P. and Xiao, Jianxiong},
title = {SUN RGB-D: A RGB-D Scene Understanding Benchmark Suite},
booktitle = {Proceedings of the IEEE Conference on Computer Vision and Pattern Recognition (CVPR)},
month = {June},
year = {2015}
}

@article{Geiger2013IJRR,
  author = {Andreas Geiger and Philip Lenz and Christoph Stiller and Raquel Urtasun},
  title = {Vision meets Robotics: The KITTI Dataset},
  journal = {International Journal of Robotics Research (IJRR)},
  year = {2013}
}

\end{document}